\documentclass{article}

\usepackage[preprint]{neurips_2023}

\usepackage[utf8]{inputenc} 
\usepackage[T1]{fontenc}    
\usepackage{hyperref}       
\usepackage{url}            
\usepackage{booktabs}       
\usepackage{amsfonts}       
\usepackage{nicefrac}       
\usepackage{microtype}      
\usepackage{xcolor}         
\usepackage{graphicx}
\usepackage{subfigure}

\title{FrameNeRF: A Simple and Efficient Framework for\\
	Few-shot Novel View Synthesis}

\author{
  Yan Xing$^1$, Pan Wang$^1$, Ligang Liu$^2$, Daolun Li$^1$, Li Zhang$^1$\\
  $^1$Hefei University of Technology\ \ \ $^2$University of Science and Technology of China\\
  \texttt{$^1$\{xingyan,2022111453,ldaol,lizhang\}@hfut.edu.cn}\ \ \   \texttt{$^2$lgliu@ustc.edu.cn}\\
}

\begin{document}

\maketitle

\begin{abstract}
We present a novel framework, called FrameNeRF, designed to apply off-the-shelf fast high-fidelity NeRF models with fast training speed and high rendering quality for few-shot novel view synthesis tasks. The training stability of fast high-fidelity models is typically constrained to dense views, making them unsuitable for few-shot novel view synthesis tasks. To address this limitation, we utilize a regularization model as a data generator to produce dense views from sparse inputs, facilitating subsequent training of fast high-fidelity models. Since these dense views are pseudo ground truth generated by the regularization model, original sparse images are then used to fine-tune the fast high-fidelity model. This process helps the model learn realistic details and correct artifacts introduced in earlier stages. By leveraging an off-the-shelf regularization model and a fast high-fidelity model, our approach achieves state-of-the-art performance across various benchmark datasets.The video is available at \href{https://github.com/wangpanpass/FrameNeRF}{https://github.com/wangpanpass/FrameNeRF}
\end{abstract}

\begin{figure}[ht]
\vskip 0.2in
\begin{center}
\centering
\includegraphics[width=1\linewidth]{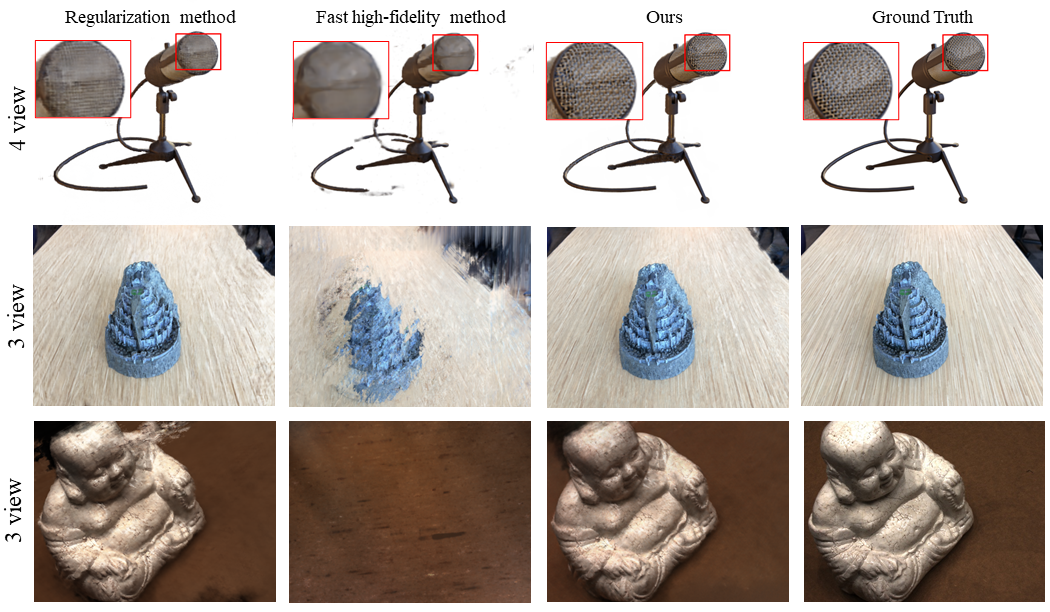}
\caption{\textbf{Example novel view synthesis results from sparse views. } The regularization method alone or the fast high-fidelity method does not perform well enough in the sparse view. Therefore, we use the regularization model and the fast high-fidelity model as components of our framework to take full advantage of their strengths and achieve a significant performance improvement in rendering quality}
\label{fig1}
\end{center}
\vskip -0.2in
\end{figure}

\begin{figure*}[ht]
\vskip 0.2in
\begin{center}
\centering
\includegraphics[width=1\linewidth]{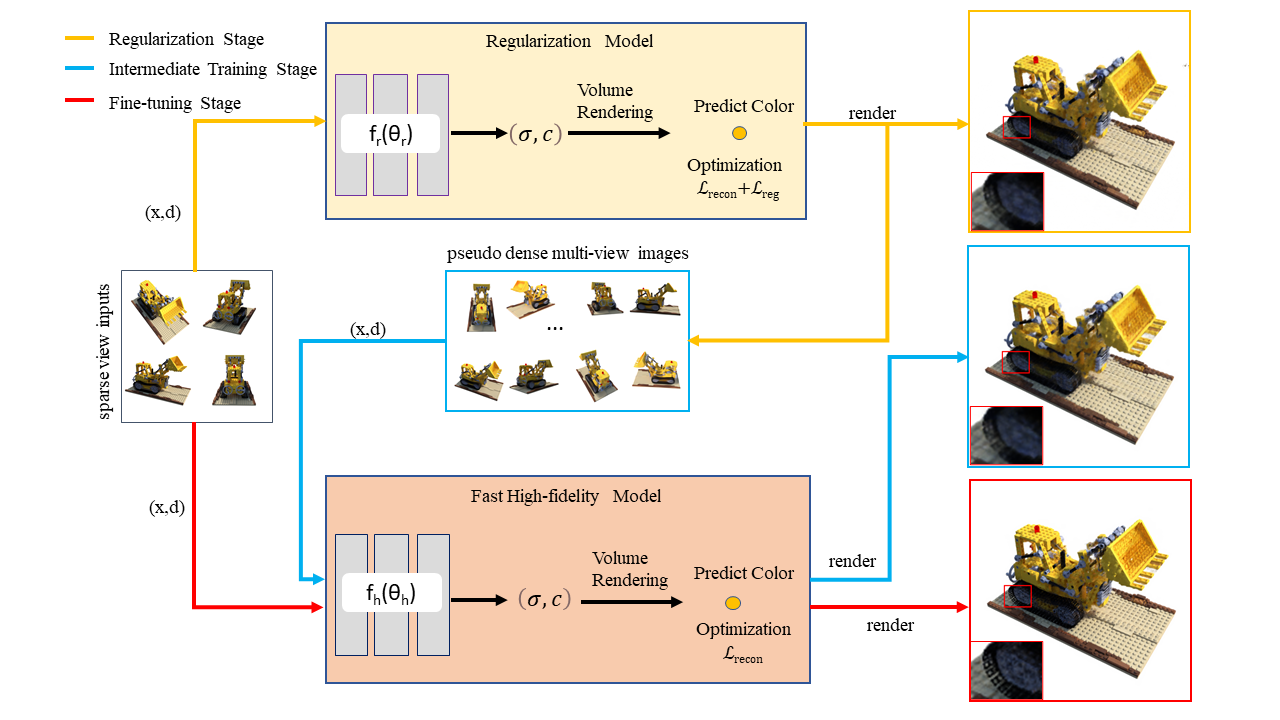}
\caption{ \textbf{Overviews of our method}. We have divided the whole training process into three stages. Regularization stage: the given sparse views are used as inputs to train the regularization model, and the dense multi-view data are generated from the trained model. Intermediate training stage: the generated pseudo-dense views are used as input to train the fast high-fidelity model. Fine-tuning stage: the original real sparse input views are used to fine-tune the high-fidelity model.
}
\label{fig2}
\end{center}
\vskip -0.2in
\end{figure*}

\section{Introduction}
\label{Introduction}
The NeRF method, based on neural networks, has achieved great success in the field of novel view synthesis due to its simplicity and realistic rendering quality. Since its release, NeRF has become one of the fundamental research paradigms in 3D vision. Many subsequent works have extended and improved NeRF, some of which have shown significant advantages in training speed and rendering quality, and we call them fast high-fidelity methods. However, the stable training of fast high-fidelity methods is limited to dense views, and when training is guided by only a limited number of views, the models are prone to overfitting to the training views, which makes it difficult to directly apply fast high-fidelity methods to the few-shot novel view synthesis task. 

Currently, the methods for solving few-shot novel view synthesis can be divided into two categories: pre-training methods and regularization methods. The pre-training approaches \cite{yu2021pixelnerf, chen2021mvsnerf,trevithick2021grf, chibane2021stereo,rematas2021sharf, johari2022geonerf, wang2021ibrnet, liu2022neural} usually pre-train on large-scale multi-view datasets to acquire geometric prior knowledge, followed by fine-tuning target scene. In contrast, regularization methods \cite{jain2021putting,roessle2022dense,wang2023sparsenerf,yang2023freenerf,deng2022depth,niemeyer2022regnerf,kim2022infonerf,kwak2023geconerf,seo2023mixnerf,seo2023flipnerf,truong2023sparf} impose various regularization constraints to ensure the stability of the training process, guiding the network to learn the correct scene geometry. Our motivation is to take full advantage of the fast high-fidelity approaches and introduce them to the few-shot novel view synthesis task. A straightforward idea is to apply the regularization settings of regularization methods to fast high-fidelity methods. However, this brings some new problems, such as the need for some modifications to the model and the fact that the validity of the model cannot be guaranteed. Therefore, we take an alternative approach and come up with a simpler and more effective solution.

In this paper, we propose FrameNeRF, which is a straightforward and effective framework for few-shot novel view synthesis. We divide the entire training process into three stages: the regularization stage, the intermediate training stage, and the fine-tuning stage. During the regularization stage, we train a regularization model from sparse views and then render dense multi-view images. The purpose of this stage is to utilize regularization models to generate pseudo-dense multi-view images, achieving data augmentation. During the intermediate training stage, we leverage the dense multi-view images obtained from the regularization model to train the fast high-fidelity model. In this stage, the fast high-fidelity model will learn a radiance field from the regularization model, which moreover improves multi-view consistency, as shown in Fig.\ref{fig3}. Since the fast high-fidelity model fits pseudo ground truth views during the intermediate training stage, this results in the model that is deficient in real scene information. Therefore, we fine-tune the fast high-fidelity model using the original sparse views in the subsequent fine-tuning stage, which further learns more details and corrects the artifacts generated in the previous stage.

In our work, the regularization model and the fast high-fidelity model serve as the two components in our framework, and FrameNeRF takes full advantage of both models to achieve better performance than a single model in the few-shot novel view synthesis task, as shown in Fig.\ref{fig1}. FrameNeRF is a simple yet effective framework that fully reuses existing models, where components can choose from different models with little or no modification to the model. Extensive experiments show that our proposed framework can reach state-of-the-art performance on multiple benchmark datasets, outperforming existing few-shot novel view synthesis methods.

\section{Related Work}
\subsection{Neural Radiance Field}
Recently, the Neural Radiance Fields (NeRF) \cite{mildenhall2021nerf} achieved impressive performance in novel view synthesis tasks. NeRF extracts the geometric and texture information of the scene from multi-view images and subsequently leverages this information to generate a Recently, the Neural Radiance Fields (NeRF) \cite{mildenhall2021nerf} achieved impressive performance in novel view synthesis tasks. NeRF extracts the geometric and texture information of the scene from multi-view images and subsequently leverages this information to generate a continuous volumetric density field, enabling the realistic synthesis of any arbitrary view. However, NeRF still has many limitations in terms of applications and performance, prompting subsequent works to extend and improve upon them. e.g. NeRF++ \cite{zhang2020nerf++} introduces inverted sphere parameterization to successfully extend the scene to unbounded 360° ones. MipNeRF\cite{barron2021mip} introduces cone-casting, emitting a cone for each pixel, which achieves anti-aliasing in multi-resolution rendering. NeRF and its variants have greatly contributed to the development of numerous research areas such as 3D reconstruction \cite{yariv2021volume, wang2021neus,wang2022hf,li2023neuralangelo}, SLAM \cite{rosinol2023nerf, avraham2022nerfels}, and 3D Generation \cite{poole2022dreamfusion, hollein2023text2room, liu2023one}, and so on. 

\subsection{Fast High-fidelity NeRFs}
Although NeRF is capable of rendering realistic novel views, the training and rendering processes are computationally intensive and time-consuming, which greatly limits the practical application. The subsequent works \cite{yu2021plenoctrees,chen2022tensorf, reiser2021kilonerf,barron2023zip,fridovich2022plenoxels,neff2021donerf,muller2022instant,hu2023tri} have improved computational efficiency, accelerating the training and rendering speed, e.g. extracting the pre-trained model into voxel grid structure \cite{yu2021plenoctrees, reiser2021kilonerf} for accelerating inference, skipping the empty and occluded areas to reduce samples along rays \cite{fridovich2022plenoxels, muller2022instant,hu2023tri}, decomposing the 4D tensor representing the radiance field into a low-rank tensor \cite{chen2022tensorf}. Some of the works exhibit significant advantages in training speed, and rendering quality, which we call fast high-fidelity methods \cite{chen2022tensorf, barron2023zip,muller2022instant,hu2023tri,chen2023neurbf}. However, fast high-fidelity methods are prone to overfitting in sparse viewpoints and cannot be directly applied to few-shot novel view synthesis tasks. Nevertheless, FrameNeRF can introduce fast high-fidelity methods to novel view synthesis tasks in a very straightforward way and fully utilize their advantages in terms of training speed and rendering quality. 

\subsection{Few-shot Novel View Synthesis}
Recently, many works have attempted to address the challenging few-shot novel view synthesis problem. These approaches can generally be categorized into two groups: pre-training methods and regularization methods. The pre-training methods obtain geometric prior knowledge by pre-training on large-scale multi-view datasets \cite{yu2021pixelnerf, chen2021mvsnerf,trevithick2021grf, chibane2021stereo, rematas2021sharf, johari2022geonerf, wang2021ibrnet, liu2022neural}, and then fine-tune the models in the target scene, e.g., CNN as a feature extractor \cite{yu2021pixelnerf, chibane2021stereo} to extract image features for training a generalizable model. Although pre-training methods have yielded promising results, pre-training datasets are costly to acquire and have a significant performance degradation for the out-of-distribution dataset. In contrast, without pre-training on large-scale multi-view datasets, regularization methods \cite{jain2021putting,roessle2022dense,wang2023sparsenerf,yang2023freenerf,deng2022depth,niemeyer2022regnerf,kim2022infonerf,kwak2023geconerf,seo2023mixnerf,seo2023flipnerf,truong2023sparf} optimize directly on the target scene by incorporating various regularization setting, e.g. depth supervision \cite{roessle2022dense, wang2023sparsenerf, deng2022depth}, cross-view semantic consistency \cite{jain2021putting}, frequency encoding regularization \cite{yang2023freenerf}, sampling unobserved viewpoints to compensate for insufficient training views \cite{niemeyer2022regnerf,kim2022infonerf,kwak2023geconerf, seo2023flipnerf}, etc. These elaborate regularizations effectively alleviate overfitting under sparse views, synthesizing desirable new views from limited observations. Our approach leverages the regularization model to generate dense multi-view images for subsequent training. The proposed FrameNeRF combines the advantages of the fast high-fidelity model and the regularization model to synthesize high-quality new perspectives from limited observations.

\section{Preliminary}
NeRF \cite{mildenhall2021nerf} represents 3D scenes as continuous volumetric density fields by neural networks with parameters $\theta$. The neural network takes the encoded coordinates $x \in R^3 $ and view direction $d \in R^2$ of the sampling point along the ray r as input, then outputs the view-independent volumetric density $\sigma \in R$ and the view-dependent color $c \in R^3$:
\begin{equation}
    \{ \gamma(x),\gamma(d) \} \longmapsto \{\sigma,c \} 
\end{equation}
where $\gamma$ denotes the positional encoding, which can map the input to a high dimensional space to facilitate the learning of high-frequency details.

NeRF \cite{mildenhall2021nerf} employs the volumetric rendering formula to denote the color of the pixel as the weighted sum of color samples along the ray $r$. The volumetric rendering formula is as follows.
\begin{equation}
    \hat{C}(r)=\sum_{i=1}^{N}{T_i(1-exp(-\sigma_i\delta_i))},\\
    T_i=exp(-\sum_{j=1}^{i-1}{\sigma_j\delta_j})
\end{equation}
where $T_i$, $c_i$, $\sigma_i$, and $\delta_i$, respectively denote the accumulated transparency, color, volume density, and distance to the next sample point for the i-th sample point. $\hat{C}(r)$ represents the estimated color of pixels. We optimize the network parameters $\theta$ by minimizing the reconstruction loss. The reconstruction loss is as follows:

\begin{equation}
    \mathcal{L}_{recon}=\sum_{r \in R}{||\hat{C}(r)-C(r)||^2}
\end{equation}
where $\mathcal{L}_{recon}$ indicates reconstruction loss, $C(r)$ denotes ground truth color for pixel, and $R$ denotes a batch of training rays. 

\section{Methodology}
Fast high-fidelity methods offer significant advantages in many aspects such as training speed and rendering quality. However, these methods cannot be directly applied to the few-shot novel view synthesis task due to aforementioned reasons such as unstable training and overfitting. Our motivation is to introduce fast high-fidelity methods to few-shot novel view synthesis tasks in a straightforward way. Fig. \ref{fig2} depicts the pipeline of our framework. Next, we will describe in detail our method, which accomplishes our goals with only a slight change in training form.

\subsection{Regularized Training on Few-shot Data}
Compared to previous methods, fast high-fidelity methods have significant advantages in terms of training speed and rendering quality. However, fast high-fidelity models are prone to overfitting to the sparse training views and the training process is unstable, thus failing to learn the correct scene geometry. To overcome the shortcomings of fast high-fidelity methods under sparse views, we can combine fast high-fidelity methods with a regularization setting in a straightforward manner. However, this introduces some new problems, e.g., it is difficult to apply frequency regularization of FreeNeRF on fast high-fidelity models (e.g., InstantNGP \cite{muller2022instant}, TensoRF \cite{chen2022tensorf}, and Tri-MipRF \cite{hu2023tri}) because these models do not use frequency encoding. Therefore, we propose an alternative solution, which is to use regularization models to generate dense multi-view images to satisfy the need of fast high-fidelity models for dense input views.

models need to learn scene geometry from sparse views, which is especially important for subsequent training. Here we choose FreeNeRF \cite{yang2023freenerf} as our regularization model. FreeNeRF \cite{yang2023freenerf} proposed frequency regularization to synchronize the convergence speed of high-frequency and low-frequency information as well as occlusion regularization to penalize near-camera artifacts so that the desirable radiance field of scene can be learned. It is worth noting that at this point the regularization model generates only coarse results that still have some artifacts and poor multi-view consistency. The regularization training process is viewed as a process of data augmentation that transforms sparse views into dense views. After obtaining dense multi-view images, we introduce existing fast high-fidelity models to few-shot novel view synthesis task with little modification in the next stage.

\subsection{Training on Dense Multi-view Data}
In the regularization stage, we use the existing regularization model as a data generator to generate dense multi-view images. Regularization In the previous stage, we obtained dense multi-view images generated by the regularization model. Since dense multi-view images ensure the stability of training as well as fully utilize the performance of the model, we utilize the obtained dense multi-view images to train the fast high-fidelity model. Considering the advantages in terms of computational efficiency and rendering quality, in our work, we choose TensoRF \cite{chen2022tensorf} (for the Blender dataset and LLFF dataset) and ZipNeRF \cite{ barron2023zip} (for the DTU dataset) as our fast high-fidelity model, respectively. In this stage, the scene knowledge learned from the regularization model is transferred to the fast high-fidelity model.

\begin{figure}[ht]
\vskip 0.2in
\begin{center}
\centering
\includegraphics[width=0.7\linewidth]{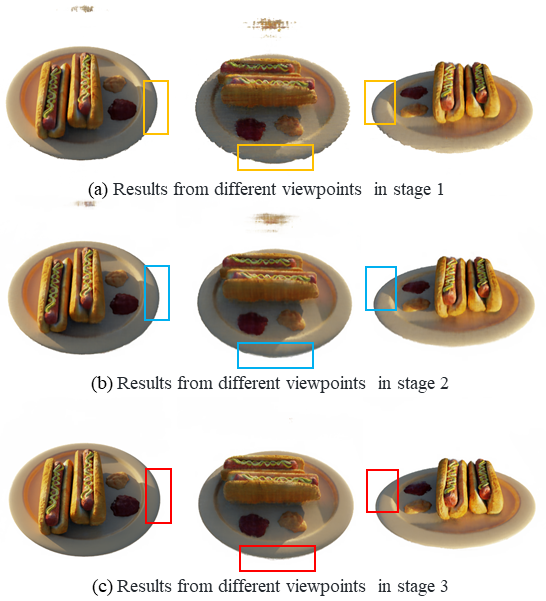}
\caption{\textbf{The rendering results in different stages from different viewpoints.}}
\label{fig3}
\end{center}
\vskip -0.2in
\end{figure}

Fig. \ref{fig3} shows the results of FrameNeRF in the different stages from different viewpoints. In Fig. \ref{fig3}(a), it can be observed that the edges of the "plate" are jagged and the same locations highlighted by the yellow boxes behave inconsistently in different viewpoints. In Fig. \ref{fig3}(b), we can observe the fast high-fidelity model yields smoother edges. This is because different views co-train the fast high-fidelity model, and the different views constrain each other, thus improving multi-view consistency. After the second stage of training, the rendering quality of the fast high-fidelity model is even slightly better than the regularization model. We can see from Fig. 3(b) that the fast high-fidelity model inherits artifacts from the regularization model in the intermediate training stage, which will be resolved in the next stage.

\subsection{Fine-tuning on Original Few-shot Input}
While regularization can guide the convergence of unsupervised regions and has shown good results in scene geometry learning, the results generated by regularization models still suffer from many floating artifacts, as shown in Fig. \ref{fig3}(a). Even more, these regularizations may prevent the model from faithfully reconstructing the scene, e.g., due to semantic regularization effects, DietNeRF \cite{jain2021putting} reconstructs 'imaginary' components that are not existent in the original scene. These dense views with artifacts produced in the first stage, which serve as inputs to the second stage, lead to undesirable effects such as floating artifacts and lack of detail in the high-fidelity model (see Fig. \ref{fig3}(b)).

\begin{figure}[ht]
\vskip 0.2in
\begin{center}
\centering
\includegraphics[width=0.7\linewidth]{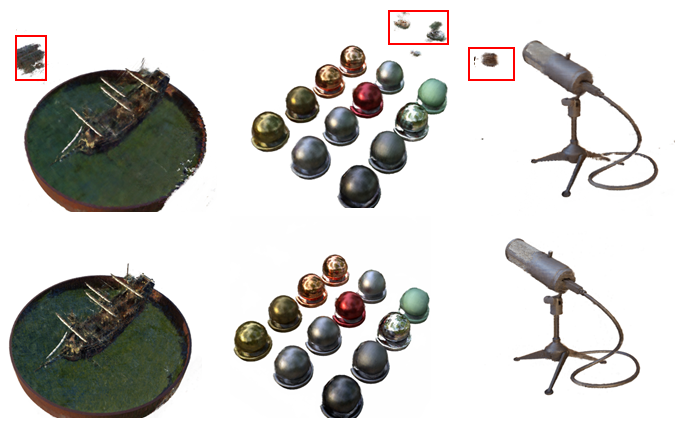}
\caption{\textbf{Impact of the fine-tuning process}. The top row shows the results with artifacts generated from the regularization model during the first stage. The bottom row depicts the results without artifacts from the fast high-fidelity model after fine-tuning.}
\label{fig4}
\end{center}
\vskip -0.2in
\end{figure}

In the fine-tuning stage, we continue to train a fast high-fidelity model using the original sparse input, taking full advantage of its advantages in rendering quality, learning more details from real sparse views, and correcting artifacts learned in previous stages. As shown in Figs. \ref{fig3}(c) and Fig. \ref{fig4}, we observe that the fine-tuned fast high-fidelity model eliminates the artifacts produced by the previous stage.

\section{Experiments}
\subsection{Experimental Settings}
\textbf{Datasets and metrics}. We evaluate the performance of our scheme on multiple benchmark datasets including Blender \cite{mildenhall2021nerf}, LLFF \cite{mildenhall2019local}, and DTU \cite{jensen2014large}. The Blender dataset consists of eight 360° bounded synthetic scenes and each scene contains 400 multi-view images. For comparison purposes, we follow the experimental protocol of FlipNeRF \cite{seo2023flipnerf}, taking the first 4/8 images of the training set as the original training inputs for stage 1 and 3 and the test set remains unchanged. For Blender, we use the 200 images predicted by the regularization model as pseudo-dense input views to stage 2. The LLFF dataset contains 8 real forward-facing scenes. We follow the protocol of RegNeRF \cite{niemeyer2022regnerf}, using every eighth image of each scene to form test set for evaluation, and uniformly sampling 3/6/9 views among the remaining images for training. The DTU data set is a large multi-view data set consisting of 124 different scenes. Like RegNeRF \cite{niemeyer2022regnerf}, we selected 15 scenes for experiments and kept the identical split of the training set and test set. For LLFF and DTU, the regularization model renders all views except the training views, and we use these generated dense views to train the fast high-fidelity model. In our work, we choose PSNR, SSIM, and LPIPS as the evaluation metrics.

\textbf{Baselines}. We compare FrameNeRF against state-of-the-art few-shot NeRF methods. We compare our method not only with regularization methods \cite{jain2021putting,wang2023sparsenerf,yang2023freenerf,niemeyer2022regnerf,truong2023sparf,seo2023mixnerf,seo2023flipnerf}but also with representative pre-training methods \cite{yu2021pixelnerf,chen2021mvsnerf,chibane2021stereo}. To ensure a fair comparison, both baselines and FrameNeRF are evaluated under the same setting. 

\textbf{Implementation details}.
In our experiments, we choose FreeNeRF \cite{yang2023freenerf} as the regularization model for all datasets, TensoRF \cite{chen2022tensorf} as the fast high-fidelity model for the Blender and LLFF datasets, and ZipNeRF \cite{barron2023zip} as the fast high-fidelity model for the DTU dataset. The experimental settings including loss functions remain almost the same as their work. The training time of the regularized model is equal to the original article, while the training time of the fast high-fidelity model (including the intermediate training stage and fine-tuning stage) is 4 minutes (for Blender), 8 minutes (for LLFF), and 14 minutes (for DTU) respectively. All experiments were done on a single RTX A40.

\subsection{ Comparisons}
We compare the novel view synthesis quality of our method with those of current state-of-the-art methods on Blender, LLFF, and DTU datasets, respectively. Extensive experiments demonstrate that FrameNeRF outperforms other methods in terms of rendering quality.

\begin{table*}[h]
\caption{\textbf{Quantitative comparisons on the Blender dataset}. Our FrameNeRF outperforms all other methods in almost all the scenarios and metrics. }
\label{tab1}
\vskip 0.15in
\begin{center}
\begin{small}
\begin{tabular}{l|cc|cc|ccr}
\toprule
          & \multicolumn{2}{c|}{PSNR$\uparrow$} & \multicolumn{2}{c|}{SSIM$\uparrow$} & \multicolumn{2}{c}{LPIPS$\downarrow$} \\
          & 4-view & 8-view & 4-view & 8-view & 4-view & 8-view \\
\midrule
DietNeRF \cite{jain2021putting}        &15.42	&21.31	&0.730	&0.879	&0.314	&0.153\\
InfoNeRF \cite{kim2022infonerf}        &18.44	&22.01	&0.792	&0.852	&0.223	&0.133\\
RegNeRF \cite{niemeyer2022regnerf}         &13.71	&19.11	&0.786	&0.841	&0.346	&0.200\\
MixNeRF \cite{seo2023mixnerf}         &18.99	&23.84	&0.807	&0.878	&0.199	&0.103\\
FlipNeRF \cite{seo2023flipnerf}        &20.60	&24.38	&0.822	&0.883	&0.159	&\textbf{0.095}\\
FreeNeRF \cite{yang2023freenerf}        &20.37	&23.72	&0.817	&0.868	&0.166	&0.115\\
FrameNeRF &\textbf{21.29}	&\textbf{25.44}	&\textbf{0.828}	&\textbf{0.887}	&\textbf{0.149}	&0.096\\
\bottomrule
\end{tabular}
\end{small}
\end{center}
\vskip -0.1in
\end{table*}

\begin{figure}[h]
\vskip 0.2in
\begin{center}
\centering
\includegraphics[width=1\linewidth]{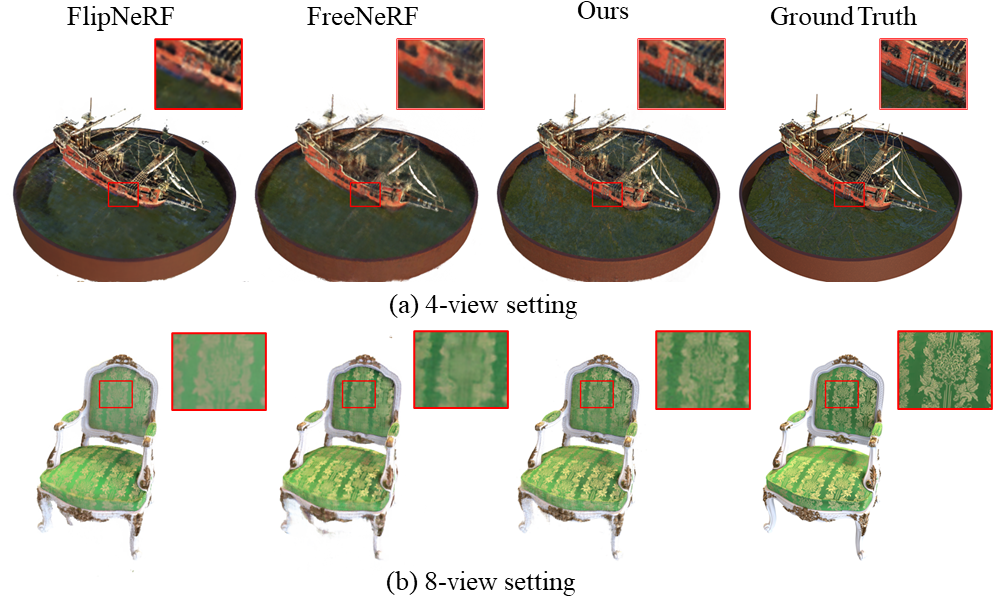}
\caption{\textbf{Qualitative comparisons on the Blender dataset}. FrameNeRF can learn more details and realistic colors.}
\label{fig5}
\end{center}
\vskip -0.2in
\end{figure}

\textbf{Blender Dataset}. Quantitative comparisons between the current state-of-the-art methods and FrameNeRF on the Blender dataset are shown in Table \ref{tab1}, where FrameNeRF outperforms all other methods. We report the reproduced results for FreeNeRF using their codebase because they chose different sparse input views. For other methods, we directly use the results reported in FlipNeRF. Fig. \ref{fig5} gives visual comparisons, in which we can observe that the chair pattern rendered by FreeNeRF is blurred, while the chair pattern rendered by FlipNeRF has distorted colors. In contrast, our FrameNeRF captures more details and renders more realistic views. We believe this is because the regularization settings of FreeNeRF and FlipNeRF affect the faithful reconstruction of the scene. In our framework, although the fast high-fidelity model learns the results from the regularization model, the fine-tuning process can correct the unrealistic artifacts learned from the previous stage and obtain more details.

\begin{table*}[h]
\caption{\textbf{Quantitative comparisons on the LLFF dataset}. On the LLFF dataset, FrameNeRF is compared with state-of-the-art regularization and pre-training methods, and achieves superior results. “ft" denotes fine-tuning.}
\label{tab2}
\vskip 0.15in
\begin{center}
\begin{small}
\begin{tabular}{l|ccc|ccc|cccr}
\toprule
          & \multicolumn{3}{c|}{PSNR$\uparrow$} & \multicolumn{3}{c|}{SSIM$\uparrow$} & \multicolumn{3}{c}{LPIPS$\downarrow$} \\
          & 3-view & 6-view & 9-view & 3-view & 6-view & 9-view & 3-view & 6-view & 9-view \\
\midrule
Pre-training &&&&&&&&&\\ \hline

PixelNeRF ft      & 16.17	& 17.03	 & 18.92   & 0.438	& 0.473	& 0.535	& 0.512	& 0.477	& 0.430 \\
SRF ft            & 17.07	& 16.75	 & 17.39   & 0.436	& 0.438	& 0.465	& 0.529	& 0.521	& 0.503 \\
MVSNeRF ft        & 17.88	& 19.99	 & 20.47   & 0.584	& 0.660	& 0.695	& 0.327	& 0.264	& 0.244 \\ \hline
Regularization &&&&&&&&\\ \hline
DietNeRF           & 14.94	& 21.75	& 24.28	& 0.370	& 0.717	& 0.801	& 0.496	& 0.248	& 0.183 \\
RegNeRF           & 19.08	& 23.10	& 24.86	& 0.587	& 0.760	& 0.820	& 0.336	& 0.206	& 0.161 \\
FreeNeRF          & 19.63	& 23.73	& 25.13	& 0.612	& 0.779	& 0.827	& 0.308	& 0.195	& 0.160 \\
SparseNeRF         & 19.86	& 23.64	& 24.97	& 0.624	& 0.784	& 0.834	& 0.328	& 0.202	& 0.158 \\ 
FrameNeRF          & \textbf{20.43}	& \textbf{24.28}	& \textbf{25.48}	&\textbf{0.669}	& \textbf{0.802}	&\textbf{0.838}	& \textbf{0.274}	&\textbf{0.186}	& \textbf{0.156} \\
\bottomrule
\end{tabular}
\end{small}
\end{center}
\vskip -0.1in
\end{table*}

%

\begin{figure}[h]
\vskip 0.2in
\begin{center}
\centering
\includegraphics[width=1\linewidth]{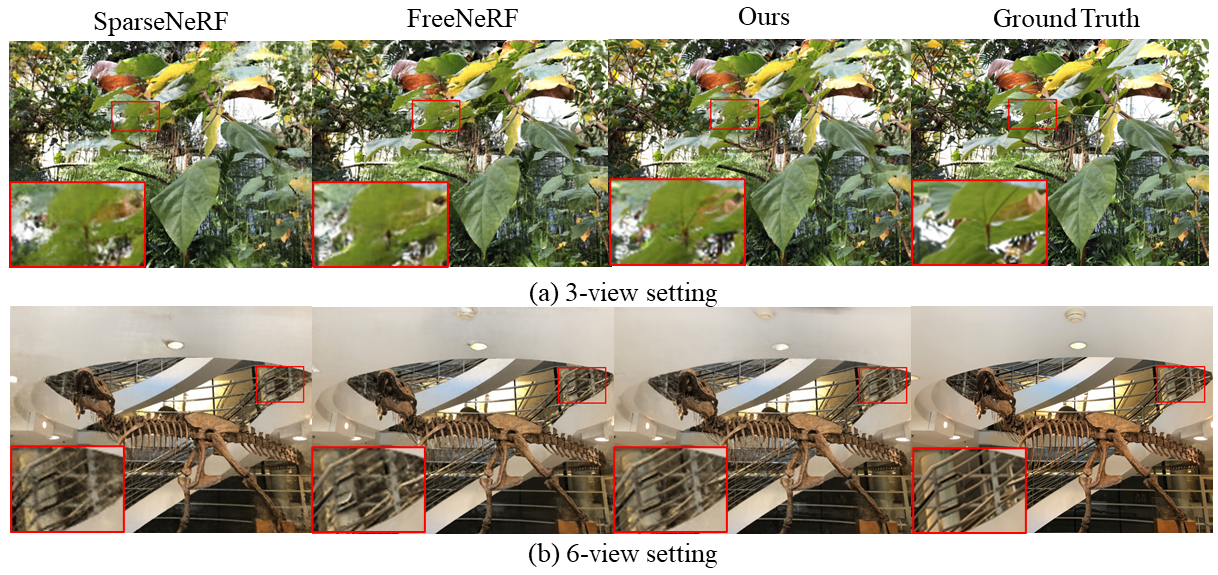}
\caption{\textbf{Qualitative comparisons on the LLFF dataset}. On the LLFF dataset, we compare with methods that perform well in sparse views. Our FrameNeRF method produces more realistic rendering results.}
\label{fig6}
\end{center}
\vskip -0.2in
\end{figure}

\textbf{LLFF Dataset}. The quantitative and qualitative results of existing state-of-the-art few-shot methods are shown in Table \ref{tab2} and Fig. \ref{fig6}, respectively. We report the reproduced results for SparseNeRF using their codebase because they lack results of 6-view and 9-view settings. For other methods, we directly use the results reported in FreeNeRF and MixNeRF. Since the pre-training methods in Table 2 are pre-trained on the DTU dataset, they show inferior performance on the LLFF dataset with large distribution differences. In contrast, regularization methods such as FreeNeRF and SparseNeRF do not require expensive pre-training to generate visually more realistic renderings. However, regularization affects the model's faithful reconstruction of the scene. It can be seen from the rendering results of FreeNeRF and SparseNeRF that leaf veins in the Leaves are missing and the railings in the Trex are blurry, as shown in Fig. \ref{fig6}. In contrast, the result of our FrameNeRF has a much finer detail.

\begin{table*}[t]
\caption{\textbf{Quantitative comparisons on DTU 3-view setting}. On the DTU dataset, our FrameNeRF outperforms all other methods, especially for the masked object image without background. "ft" denotes fine-tuning.}
\label{tab3}
\vskip 0.15in
\begin{center}
\begin{small}
\begin{tabular}{l|ccc|cccr}
\toprule
          & \multicolumn{3}{c|}{Object image} & \multicolumn{3}{c}{Full images}  \\
          & PSNR$\uparrow$ & SSIM$\uparrow$ & LPIPS$\downarrow$ & PSNR$\uparrow$ & SSIM$\uparrow$ & LPIPS$\downarrow$ \\
\midrule
SRF \cite{chibane2021stereo} ft	        &15.68	&0.698	&0.281	&16.06	&0.550	&0.431\\
PixelNeRF \cite{yu2021pixelnerf} ft 	&18.95	&0.710	&0.269	&17.38	&0.548	&0.456\\
MVSNeRF \cite{chen2021mvsnerf} ft	    &18.54	&0.769	&0.197	&16.26	&0.601	&0.384\\
MixNeRF \cite{seo2023mixnerf}	        &18.95	&0.744	&0.203	&16.05	&0.606	&0.367\\
FlipNeRF \cite{seo2023flipnerf}	    &19.55	&0.767	&0.180	&15.70	&0.593	&0.359\\
SparseNeRF \cite{truong2023sparf}	    &19.47	&0.829	&0.183	&-	    &-	    &-\\
FreeNeRF \cite{yang2023freenerf}	    &19.63	&0.612	&0.182	&18.02	&0.680	&0.318\\
FrameNeRF	    &\textbf{24.15}	&\textbf{0.834}	&\textbf{0.072}	&\textbf{18.64}	&\textbf{0.709}	&\textbf{0.182}\\
\bottomrule
\end{tabular}
\end{small}
\end{center}
\vskip -0.1in
\end{table*}

\begin{figure*}[h]
\vskip 0.2in
\begin{center}
\centering
\includegraphics[width=1\linewidth]{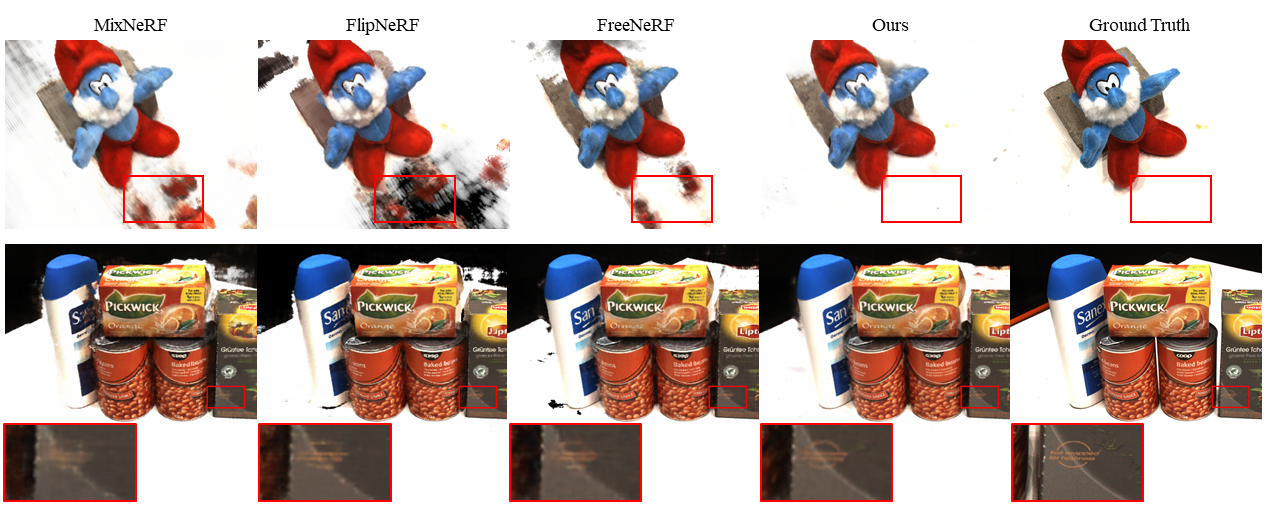}
\caption{\textbf{Qualitative comparisons on the DTU dataset}. Our FrameNeRF can render new views with fewer artifacts and sharper details.}
\label{fig7}
\end{center}
\vskip -0.2in
\end{figure*}

\textbf{DTU Dataset}.Table\ref{tab3} shows the quantitative results of FrameNeRF and baselines on the DTU dataset for full image and masked object image. We evaluate full image metrics for MixNeRF and FlipNeRF by using their trained models. For other methods, we directly use the results reported in FreeNeRF. From Table\ref{tab3} we can see that our method outperforms baseline methods, especially for object image metrics in the 3-view setting. Fig. \ref{fig7} shows that the results of our method have fewer artifacts and sharper details compared to the baseline method.

\subsection{Ablation Study}
To verify the rationality and validity of our design, we ablate the training stage and model choices in our framework.

\textbf{Ablation of training stage}. Fig.\ref{fig3}(a) shows that the rendering result of the regularization stage contains artifacts and the plate has jagged edges, Fig.\ref{fig3}(b) observes that the plate edges become smoother after the intermediate training stage, and Fig.\ref{fig3}(c) shows that the fine-tuning stage removes the original artifacts and the rendering result is more realistic. This suggests that both the intermediate training and fine-tuning stages further improve rendering quality compared to the previous stage. As shown in Table \ref{tab4}, we show quantitative results for different training stages on multiple datasets. The above qualitative and quantitative results validate the effectiveness of the three-stage design.

\begin{table}[t]
\caption{\textbf{Quantitative PSNR comparisons of outcomes at different training stages}.}
\label{tab4}
\vskip 0.15in
\begin{center}
\begin{small}
\begin{tabular}{lc|ccc}
\toprule
 &&       \multicolumn{3}{c}{Training stage}  \\
Dataset &view & stage 1 &  stage 2 & stage 3  \\
\midrule
Blender	&4	&20.37	&20.86	&22.14  \\
LLFF	&3	&19.63	&19.77	&20.43  \\
DTU	    &3	&18.01  &18.25	&18.64  \\
\bottomrule
\end{tabular}
\end{small}
\end{center}
\vskip -0.1in
\end{table}

\textbf{Ablation of submodules}. Taking the hotdog of Blender as an example, Table \ref{tab5} shows the impact of different choices of regularization models and fast high-fidelity models. In our framework, we can flexibly choose sub-modules, and we combine various candidate regularization models (FreeNeRF, MixNeRF, FlipNeRF) and fast high-fidelity models (TensoRF, NeuRBF, and InstantNGP) to select the best-performing scheme. In Table \ref{tab5}, we take the hotdog scenario of the Blender dataset as an example to search for the optimal combination of the regularization model and fast high-fidelity model. Upon comparing the rows in the table, FreeNeRF emerges as the best choice for the regularization model. When comparing the columns in the table, TensoRF stands out as the optimal choice. Therefore, we opt for the combination of FreeNeRF and TensoRF as the solution for most of the datasets. However, for the DTU dataset with background, TensoRF fails to adequately model the background. Consequently, we select the combination of FreeNeRF and ZipNeRF as the solution. 

\begin{figure}[h]
\vskip 0.2in
\begin{center}
\centering
\includegraphics[width=1\linewidth]{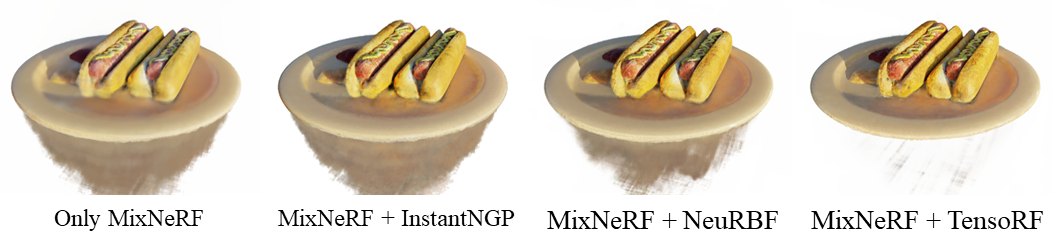}
\caption{\textbf{Impact of different fast high-fidelity models}. TensoRF\cite{chen2022tensorf} as the fast high-fidelity model can remove occlusion artifacts.}
\label{fig8}
\end{center}
\vskip -0.2in
\end{figure}

MixNeRF introduces occlusion artifacts when used as the regularization model. As can be seen from the last column of Table \ref{tab5}, the results of the combination scheme of MixNeRF plus TensoRF are significantly better than the other combinations. This is because TensoRF as the high-fidelity model can remove occlusion artifacts more effectively than NeuRBF and InstantNGP, as shown in Fig. \ref{fig8}.

\begin{table}[t]
\caption{\textbf{PSNR results for different combinations of submodules in the hotdog scene}. Here, the numbers before and after the "/"; represent the PSNR of the newly synthesized views under 4-view and 8-view input settings respectively. “Base" denotes the standalone regularization model.}
\label{tab5}
\vskip 0.15in
\begin{center}
\begin{small}
\begin{tabular}{l|ccc}
\toprule
Combination    &FreeNeRF &FlipNeRF &MixNeRF \\
\midrule 
Base                &25.55/28.00	&25.70/27.84	&16.16/25.28  \\
InstantNGP	&27.15/\textbf{30.47}	&25.79/28.53	&16.22/27.96  \\
NeuRBF        &27.63/30.33	&26.24/29.08	&17.95/28.45  \\
TensoRF        &\textbf{27.94}/29.50	&\textbf{27.20}/\textbf{29.12}	&\textbf{21.96}/\textbf{28.80}  \\
\bottomrule
\end{tabular}
\end{small}
\end{center}
\vskip -0.1in
\end{table}

\section{Conclusion}
We introduce FrameNeRF, a simple and powerful training framework for few-shot novel view synthesis. Our approach leverages existing regularization models and high-fidelity models as components of our training framework and successfully synthesizes high-quality novel views from limited observations through three stages of training. FrameNeRF outperforms existing state-of-the-art methods on multiple benchmark datasets. Our framework can fully reuse existing SOTA modules, meanwhile, our framework can evolve over time and be updated with more advanced and appropriate modules to further improve performance. The choice of modules is flexible. For example, the regularization model can be replaced with faster ZeroRF \cite{shi2023zerorf}, which can significantly reduce the overall training time. State-of-the-art 3D reconstruction methods \cite{wu2022voxurf,li2023neuralangelo}, serving as the high-fidelity model, probably achieve promising surface reconstruction from sparse input. We hope that our work will inspire more thinking about few-shot tasks.

\bibliography{Preprint}

\begin{thebibliography}{44}
\providecommand{\natexlab}[1]{#1}
\providecommand{\url}[1]{\texttt{#1}}
\expandafter\ifx\csname urlstyle\endcsname\relax
  \providecommand{\doi}[1]{doi: #1}\else
  \providecommand{\doi}{doi: \begingroup \urlstyle{rm}\Url}\fi

\bibitem[Avraham et~al.(2022)Avraham, Straub, Shen, Yang, Germain, Sweeney,
  Balntas, Novotny, DeTone, and Newcombe]{avraham2022nerfels}
Avraham, G., Straub, J., Shen, T., Yang, T.-Y., Germain, H., Sweeney, C.,
  Balntas, V., Novotny, D., DeTone, D., and Newcombe, R.
\newblock Nerfels: renderable neural codes for improved camera pose estimation.
\newblock In \emph{Proceedings of the IEEE/CVF Conference on Computer Vision
  and Pattern Recognition}, pp.\  5061--5070, 2022.

\bibitem[Barron et~al.(2021)Barron, Mildenhall, Tancik, Hedman, Martin-Brualla,
  and Srinivasan]{barron2021mip}
Barron, J.~T., Mildenhall, B., Tancik, M., Hedman, P., Martin-Brualla, R., and
  Srinivasan, P.~P.
\newblock Mip-nerf: A multiscale representation for anti-aliasing neural
  radiance fields.
\newblock In \emph{Proceedings of the IEEE/CVF International Conference on
  Computer Vision}, pp.\  5855--5864, 2021.

\bibitem[Barron et~al.(2023)Barron, Mildenhall, Verbin, Srinivasan, and
  Hedman]{barron2023zip}
Barron, J.~T., Mildenhall, B., Verbin, D., Srinivasan, P.~P., and Hedman, P.
\newblock Zip-nerf: Anti-aliased grid-based neural radiance fields.
\newblock \emph{arXiv preprint arXiv:2304.06706}, 2023.

\bibitem[Chen et~al.(2021)Chen, Xu, Zhao, Zhang, Xiang, Yu, and
  Su]{chen2021mvsnerf}
Chen, A., Xu, Z., Zhao, F., Zhang, X., Xiang, F., Yu, J., and Su, H.
\newblock Mvsnerf: Fast generalizable radiance field reconstruction from
  multi-view stereo.
\newblock In \emph{Proceedings of the IEEE/CVF International Conference on
  Computer Vision}, pp.\  14124--14133, 2021.

\bibitem[Chen et~al.(2022)Chen, Xu, Geiger, Yu, and Su]{chen2022tensorf}
Chen, A., Xu, Z., Geiger, A., Yu, J., and Su, H.
\newblock Tensorf: Tensorial radiance fields.
\newblock In \emph{European Conference on Computer Vision}, pp.\  333--350.
  Springer, 2022.

\bibitem[Chen et~al.(2023)Chen, Li, Song, Chen, Yu, Yuan, and
  Xu]{chen2023neurbf}
Chen, Z., Li, Z., Song, L., Chen, L., Yu, J., Yuan, J., and Xu, Y.
\newblock Neurbf: A neural fields representation with adaptive radial basis
  functions.
\newblock In \emph{Proceedings of the IEEE/CVF International Conference on
  Computer Vision}, pp.\  4182--4194, 2023.

\bibitem[Chibane et~al.(2021)Chibane, Bansal, Lazova, and
  Pons-Moll]{chibane2021stereo}
Chibane, J., Bansal, A., Lazova, V., and Pons-Moll, G.
\newblock Stereo radiance fields (srf): Learning view synthesis for sparse
  views of novel scenes.
\newblock In \emph{Proceedings of the IEEE/CVF Conference on Computer Vision
  and Pattern Recognition}, pp.\  7911--7920, 2021.

\bibitem[Deng et~al.(2022)Deng, Liu, Zhu, and Ramanan]{deng2022depth}
Deng, K., Liu, A., Zhu, J.-Y., and Ramanan, D.
\newblock Depth-supervised nerf: Fewer views and faster training for free.
\newblock In \emph{Proceedings of the IEEE/CVF Conference on Computer Vision
  and Pattern Recognition}, pp.\  12882--12891, 2022.

\bibitem[Fridovich-Keil et~al.(2022)Fridovich-Keil, Yu, Tancik, Chen, Recht,
  and Kanazawa]{fridovich2022plenoxels}
Fridovich-Keil, S., Yu, A., Tancik, M., Chen, Q., Recht, B., and Kanazawa, A.
\newblock Plenoxels: Radiance fields without neural networks.
\newblock In \emph{Proceedings of the IEEE/CVF Conference on Computer Vision
  and Pattern Recognition}, pp.\  5501--5510, 2022.

\bibitem[H{\"o}llein et~al.(2023)H{\"o}llein, Cao, Owens, Johnson, and
  Nie{\ss}ner]{hollein2023text2room}
H{\"o}llein, L., Cao, A., Owens, A., Johnson, J., and Nie{\ss}ner, M.
\newblock Text2room: Extracting textured 3d meshes from 2d text-to-image
  models.
\newblock \emph{arXiv preprint arXiv:2303.11989}, 2023.

\bibitem[Hu et~al.(2023)Hu, Wang, Ma, Yang, Gao, Liu, and Ma]{hu2023tri}
Hu, W., Wang, Y., Ma, L., Yang, B., Gao, L., Liu, X., and Ma, Y.
\newblock Tri-miprf: Tri-mip representation for efficient anti-aliasing neural
  radiance fields.
\newblock In \emph{Proceedings of the IEEE/CVF International Conference on
  Computer Vision}, pp.\  19774--19783, 2023.

\bibitem[Jain et~al.(2021)Jain, Tancik, and Abbeel]{jain2021putting}
Jain, A., Tancik, M., and Abbeel, P.
\newblock Putting nerf on a diet: Semantically consistent few-shot view
  synthesis.
\newblock In \emph{Proceedings of the IEEE/CVF International Conference on
  Computer Vision}, pp.\  5885--5894, 2021.

\bibitem[Jensen et~al.(2014)Jensen, Dahl, Vogiatzis, Tola, and
  Aan{\ae}s]{jensen2014large}
Jensen, R., Dahl, A., Vogiatzis, G., Tola, E., and Aan{\ae}s, H.
\newblock Large scale multi-view stereopsis evaluation.
\newblock In \emph{Proceedings of the IEEE conference on computer vision and
  pattern recognition}, pp.\  406--413, 2014.

\bibitem[Johari et~al.(2022)Johari, Lepoittevin, and
  Fleuret]{johari2022geonerf}
Johari, M.~M., Lepoittevin, Y., and Fleuret, F.
\newblock Geonerf: Generalizing nerf with geometry priors.
\newblock In \emph{Proceedings of the IEEE/CVF Conference on Computer Vision
  and Pattern Recognition}, pp.\  18365--18375, 2022.

\bibitem[Kim et~al.(2022)Kim, Seo, and Han]{kim2022infonerf}
Kim, M., Seo, S., and Han, B.
\newblock Infonerf: Ray entropy minimization for few-shot neural volume
  rendering.
\newblock In \emph{Proceedings of the IEEE/CVF Conference on Computer Vision
  and Pattern Recognition}, pp.\  12912--12921, 2022.

\bibitem[Kwak et~al.(2023)Kwak, Song, and Kim]{kwak2023geconerf}
Kwak, M., Song, J., and Kim, S.
\newblock Geconerf: Few-shot neural radiance fields via geometric consistency.
\newblock \emph{arXiv preprint arXiv:2301.10941}, 2023.

\bibitem[Li et~al.(2023)Li, M{\"u}ller, Evans, Taylor, Unberath, Liu, and
  Lin]{li2023neuralangelo}
Li, Z., M{\"u}ller, T., Evans, A., Taylor, R.~H., Unberath, M., Liu, M.-Y., and
  Lin, C.-H.
\newblock Neuralangelo: High-fidelity neural surface reconstruction.
\newblock In \emph{Proceedings of the IEEE/CVF Conference on Computer Vision
  and Pattern Recognition}, pp.\  8456--8465, 2023.

\bibitem[Liu et~al.(2023)Liu, Xu, Jin, Chen, Xu, Su, et~al.]{liu2023one}
Liu, M., Xu, C., Jin, H., Chen, L., Xu, Z., Su, H., et~al.
\newblock One-2-3-45: Any single image to 3d mesh in 45 seconds without
  per-shape optimization.
\newblock \emph{arXiv preprint arXiv:2306.16928}, 2023.

\bibitem[Liu et~al.(2022)Liu, Peng, Liu, Wang, Wang, Theobalt, Zhou, and
  Wang]{liu2022neural}
Liu, Y., Peng, S., Liu, L., Wang, Q., Wang, P., Theobalt, C., Zhou, X., and
  Wang, W.
\newblock Neural rays for occlusion-aware image-based rendering.
\newblock In \emph{Proceedings of the IEEE/CVF Conference on Computer Vision
  and Pattern Recognition}, pp.\  7824--7833, 2022.

\bibitem[Mildenhall et~al.(2019)Mildenhall, Srinivasan, Ortiz-Cayon, Kalantari,
  Ramamoorthi, Ng, and Kar]{mildenhall2019local}
Mildenhall, B., Srinivasan, P.~P., Ortiz-Cayon, R., Kalantari, N.~K.,
  Ramamoorthi, R., Ng, R., and Kar, A.
\newblock Local light field fusion: Practical view synthesis with prescriptive
  sampling guidelines.
\newblock \emph{ACM Transactions on Graphics (TOG)}, 38\penalty0 (4):\penalty0
  1--14, 2019.

\bibitem[Mildenhall et~al.(2021)Mildenhall, Srinivasan, Tancik, Barron,
  Ramamoorthi, and Ng]{mildenhall2021nerf}
Mildenhall, B., Srinivasan, P.~P., Tancik, M., Barron, J.~T., Ramamoorthi, R.,
  and Ng, R.
\newblock Nerf: Representing scenes as neural radiance fields for view
  synthesis.
\newblock \emph{Communications of the ACM}, 65\penalty0 (1):\penalty0 99--106,
  2021.

\bibitem[M{\"u}ller et~al.(2022)M{\"u}ller, Evans, Schied, and
  Keller]{muller2022instant}
M{\"u}ller, T., Evans, A., Schied, C., and Keller, A.
\newblock Instant neural graphics primitives with a multiresolution hash
  encoding.
\newblock \emph{ACM Transactions on Graphics (ToG)}, 41\penalty0 (4):\penalty0
  1--15, 2022.

\bibitem[Neff et~al.(2021)Neff, Stadlbauer, Parger, Kurz, Mueller, Chaitanya,
  Kaplanyan, and Steinberger]{neff2021donerf}
Neff, T., Stadlbauer, P., Parger, M., Kurz, A., Mueller, J.~H., Chaitanya, C.
  R.~A., Kaplanyan, A., and Steinberger, M.
\newblock Donerf: Towards real-time rendering of compact neural radiance fields
  using depth oracle networks.
\newblock In \emph{Computer Graphics Forum}, volume~40, pp.\  45--59. Wiley
  Online Library, 2021.

\bibitem[Niemeyer et~al.(2022)Niemeyer, Barron, Mildenhall, Sajjadi, Geiger,
  and Radwan]{niemeyer2022regnerf}
Niemeyer, M., Barron, J.~T., Mildenhall, B., Sajjadi, M.~S., Geiger, A., and
  Radwan, N.
\newblock Regnerf: Regularizing neural radiance fields for view synthesis from
  sparse inputs.
\newblock In \emph{Proceedings of the IEEE/CVF Conference on Computer Vision
  and Pattern Recognition}, pp.\  5480--5490, 2022.

\bibitem[Poole et~al.(2022)Poole, Jain, Barron, and
  Mildenhall]{poole2022dreamfusion}
Poole, B., Jain, A., Barron, J.~T., and Mildenhall, B.
\newblock Dreamfusion: Text-to-3d using 2d diffusion.
\newblock \emph{arXiv preprint arXiv:2209.14988}, 2022.

\bibitem[Reiser et~al.(2021)Reiser, Peng, Liao, and Geiger]{reiser2021kilonerf}
Reiser, C., Peng, S., Liao, Y., and Geiger, A.
\newblock Kilonerf: Speeding up neural radiance fields with thousands of tiny
  mlps.
\newblock In \emph{Proceedings of the IEEE/CVF International Conference on
  Computer Vision}, pp.\  14335--14345, 2021.

\bibitem[Rematas et~al.(2021)Rematas, Martin-Brualla, and
  Ferrari]{rematas2021sharf}
Rematas, K., Martin-Brualla, R., and Ferrari, V.
\newblock Sharf: Shape-conditioned radiance fields from a single view.
\newblock \emph{arXiv preprint arXiv:2102.08860}, 2021.

\bibitem[Roessle et~al.(2022)Roessle, Barron, Mildenhall, Srinivasan, and
  Nie{\ss}ner]{roessle2022dense}
Roessle, B., Barron, J.~T., Mildenhall, B., Srinivasan, P.~P., and Nie{\ss}ner,
  M.
\newblock Dense depth priors for neural radiance fields from sparse input
  views.
\newblock In \emph{Proceedings of the IEEE/CVF Conference on Computer Vision
  and Pattern Recognition}, pp.\  12892--12901, 2022.

\bibitem[Rosinol et~al.(2023)Rosinol, Leonard, and Carlone]{rosinol2023nerf}
Rosinol, A., Leonard, J.~J., and Carlone, L.
\newblock Nerf-slam: Real-time dense monocular slam with neural radiance
  fields.
\newblock In \emph{2023 IEEE/RSJ International Conference on Intelligent Robots
  and Systems (IROS)}, pp.\  3437--3444. IEEE, 2023.

\bibitem[Seo et~al.(2023{\natexlab{a}})Seo, Chang, and Kwak]{seo2023flipnerf}
Seo, S., Chang, Y., and Kwak, N.
\newblock Flipnerf: Flipped reflection rays for few-shot novel view synthesis.
\newblock In \emph{Proceedings of the IEEE/CVF International Conference on
  Computer Vision}, pp.\  22883--22893, 2023{\natexlab{a}}.

\bibitem[Seo et~al.(2023{\natexlab{b}})Seo, Han, Chang, and
  Kwak]{seo2023mixnerf}
Seo, S., Han, D., Chang, Y., and Kwak, N.
\newblock Mixnerf: Modeling a ray with mixture density for novel view synthesis
  from sparse inputs.
\newblock In \emph{Proceedings of the IEEE/CVF Conference on Computer Vision
  and Pattern Recognition}, pp.\  20659--20668, 2023{\natexlab{b}}.

\bibitem[Shi et~al.(2023)Shi, Wei, Wang, and Su]{shi2023zerorf}
Shi, R., Wei, X., Wang, C., and Su, H.
\newblock Zerorf: Fast sparse view 360 $\{$$\backslash$deg$\}$ reconstruction
  with zero pretraining.
\newblock \emph{arXiv preprint arXiv:2312.09249}, 2023.

\bibitem[Trevithick \& Yang(2021)Trevithick and Yang]{trevithick2021grf}
Trevithick, A. and Yang, B.
\newblock Grf: Learning a general radiance field for 3d representation and
  rendering.
\newblock In \emph{Proceedings of the IEEE/CVF International Conference on
  Computer Vision}, pp.\  15182--15192, 2021.

\bibitem[Truong et~al.(2023)Truong, Rakotosaona, Manhardt, and
  Tombari]{truong2023sparf}
Truong, P., Rakotosaona, M.-J., Manhardt, F., and Tombari, F.
\newblock Sparf: Neural radiance fields from sparse and noisy poses.
\newblock In \emph{Proceedings of the IEEE/CVF Conference on Computer Vision
  and Pattern Recognition}, pp.\  4190--4200, 2023.

\bibitem[Wang et~al.(2023)Wang, Chen, Loy, and Liu]{wang2023sparsenerf}
Wang, G., Chen, Z., Loy, C.~C., and Liu, Z.
\newblock Sparsenerf: Distilling depth ranking for few-shot novel view
  synthesis.
\newblock \emph{arXiv preprint arXiv:2303.16196}, 2023.

\bibitem[Wang et~al.(2021{\natexlab{a}})Wang, Liu, Liu, Theobalt, Komura, and
  Wang]{wang2021neus}
Wang, P., Liu, L., Liu, Y., Theobalt, C., Komura, T., and Wang, W.
\newblock Neus: Learning neural implicit surfaces by volume rendering for
  multi-view reconstruction.
\newblock \emph{arXiv preprint arXiv:2106.10689}, 2021{\natexlab{a}}.

\bibitem[Wang et~al.(2021{\natexlab{b}})Wang, Wang, Genova, Srinivasan, Zhou,
  Barron, Martin-Brualla, Snavely, and Funkhouser]{wang2021ibrnet}
Wang, Q., Wang, Z., Genova, K., Srinivasan, P.~P., Zhou, H., Barron, J.~T.,
  Martin-Brualla, R., Snavely, N., and Funkhouser, T.
\newblock Ibrnet: Learning multi-view image-based rendering.
\newblock In \emph{Proceedings of the IEEE/CVF Conference on Computer Vision
  and Pattern Recognition}, pp.\  4690--4699, 2021{\natexlab{b}}.

\bibitem[Wang et~al.(2022)Wang, Skorokhodov, and Wonka]{wang2022hf}
Wang, Y., Skorokhodov, I., and Wonka, P.
\newblock Hf-neus: Improved surface reconstruction using high-frequency
  details.
\newblock \emph{Advances in Neural Information Processing Systems},
  35:\penalty0 1966--1978, 2022.

\bibitem[Wu et~al.(2022)Wu, Wang, Pan, Xu, Theobalt, Liu, and
  Lin]{wu2022voxurf}
Wu, T., Wang, J., Pan, X., Xu, X., Theobalt, C., Liu, Z., and Lin, D.
\newblock Voxurf: Voxel-based efficient and accurate neural surface
  reconstruction.
\newblock \emph{arXiv preprint arXiv:2208.12697}, 2022.

\bibitem[Yang et~al.(2023)Yang, Pavone, and Wang]{yang2023freenerf}
Yang, J., Pavone, M., and Wang, Y.
\newblock Freenerf: Improving few-shot neural rendering with free frequency
  regularization.
\newblock In \emph{Proceedings of the IEEE/CVF Conference on Computer Vision
  and Pattern Recognition}, pp.\  8254--8263, 2023.

\bibitem[Yariv et~al.(2021)Yariv, Gu, Kasten, and Lipman]{yariv2021volume}
Yariv, L., Gu, J., Kasten, Y., and Lipman, Y.
\newblock Volume rendering of neural implicit surfaces.
\newblock \emph{Advances in Neural Information Processing Systems},
  34:\penalty0 4805--4815, 2021.

\bibitem[Yu et~al.(2021{\natexlab{a}})Yu, Li, Tancik, Li, Ng, and
  Kanazawa]{yu2021plenoctrees}
Yu, A., Li, R., Tancik, M., Li, H., Ng, R., and Kanazawa, A.
\newblock Plenoctrees for real-time rendering of neural radiance fields.
\newblock In \emph{Proceedings of the IEEE/CVF International Conference on
  Computer Vision}, pp.\  5752--5761, 2021{\natexlab{a}}.

\bibitem[Yu et~al.(2021{\natexlab{b}})Yu, Ye, Tancik, and
  Kanazawa]{yu2021pixelnerf}
Yu, A., Ye, V., Tancik, M., and Kanazawa, A.
\newblock pixelnerf: Neural radiance fields from one or few images.
\newblock In \emph{Proceedings of the IEEE/CVF Conference on Computer Vision
  and Pattern Recognition}, pp.\  4578--4587, 2021{\natexlab{b}}.

\bibitem[Zhang et~al.(2020)Zhang, Riegler, Snavely, and
  Koltun]{zhang2020nerf++}
Zhang, K., Riegler, G., Snavely, N., and Koltun, V.
\newblock Nerf++: Analyzing and improving neural radiance fields.
\newblock \emph{arXiv preprint arXiv:2010.07492}, 2020.

\end{thebibliography}
\bibliographystyle{icml2024}

\end{document}